\documentclass[letterpaper]{article} 
\usepackage{aaai2026}  
\usepackage{times}  
\usepackage{helvet}  
\usepackage{courier}  
\usepackage[hyphens]{url}  
\usepackage{graphicx} 
\usepackage{natbib}  
\usepackage{caption} 
\nocopyright
\usepackage{algorithm}
\usepackage{algorithmic}
\usepackage{booktabs}
\usepackage{listings} 
\usepackage{subcaption}
\usepackage{ragged2e}
\usepackage{newfloat}
\usepackage{listings}
\DeclareCaptionStyle{ruled}{labelfont=normalfont,labelsep=colon,strut=off} 
\floatstyle{ruled}
\newfloat{listing}{tb}{lst}{}
\floatname{listing}{Listing}
\title{HEALTH-PARIKSHA: Assessing RAG Models for Health Chatbots in Real-World Multilingual Settings}
\author{
    Varun Gumma\textsuperscript{\rm 1}\thanks{Work done at Microsoft},
    Ananditha Raghunath\textsuperscript{\rm 2}\footnotemark[1], 
    Mohit Jain\textsuperscript{\rm 3}\equalcontrib,
    Sunayana Sitaram\textsuperscript{\rm 3}\equalcontrib
}
\affiliations{
    \textsuperscript{\rm 1}Nanyang Technological University, \\
    \textsuperscript{\rm 2}University of Washington, \\
    \textsuperscript{\rm 3}Microsoft Corporation\\

    varun.gumma@protonmail.com, sunayana.sitaram@microsoft.com
}

\newcommand{\factualcorrectness}{\textsc{Factual Correctness}}
\newcommand{\semanticsimilarity}{\textsc{Semantic Similarity}}
\newcommand{\coherence}{\textsc{Coherence}}
\newcommand{\conciseness}{\textsc{Conciseness}}

\newcommand{\aggregate}{\textsc{Aggregate}}
\newcommand{\redactedname}{\textsc{Karya}}
\newcommand{\fturl}[1]{\footnote{\url{#1}}}
\definecolor{mygreen}{HTML}{d9ead3}
\definecolor{myorange}{HTML}{fce5cd}
\definecolor{myyellow}{HTML}{fff2cc}
\definecolor{myred}{HTML}{f4cccc}
\definecolor{mymagenta}{HTML}{fff2cc}
\definecolor{myblue}{HTML}{cfe2f3}
\definecolor{mygray}{HTML}{efefef}

\begin{document}

\maketitle

\begin{abstract}
Assessing the capabilities and limitations of Large Language Models has garnered significant interest, yet the evaluation of multiple models in real-world scenarios remains rare. Multilingual evaluation often relies on translated benchmarks, which typically do not capture linguistic and cultural nuances present in the source language. This study provides an extensive assessment of 24 LLMs on real-world data collected from Indian patients interacting with a medical chatbot in Indian English and 4 other Indic languages. We employ a uniform Retrieval Augmented Generation framework to generate responses, which are evaluated using both automated techniques and human evaluators on four specific metrics relevant to our application. We find that models vary significantly in their performance and that instruction-tuned Indic models do not always perform well on Indic language queries. Further, we empirically show that factual correctness is generally lower for responses to Indic queries compared to English queries. Finally, our qualitative work shows that code-mixed and culturally relevant queries in our dataset pose challenges to evaluated models. 
\end{abstract}


\section{Introduction}
\label{sec:introduction}
Large Language Models (LLMs) have demonstrated impressive proficiency across various domains. Nonetheless, their full spectrum of capabilities and limitations remains unclear, resulting in unpredictable performance on certain tasks. Additionally, there is now a wide selection of LLMs available. Therefore, evaluation has become crucial for comprehending the internal mechanisms of LLMs and for comparing them against each other. 

Despite the importance of evaluation, significant challenges persist. Many widely-used benchmarks for assessing LLMs are contaminated \cite{ahuja-etal-2024-megaverse,oren2024proving,xu2024benchmark,deng-etal-2024-investigating}, meaning that they often appear in LLM training data. Some of these benchmarks were created for conventional Natural Language Processing tasks and may not fully represent current practical applications of LLMs \cite{conneau-etal-2018-xnli,pan-etal-2017-cross}. Recently, there has been growing interest in assessing LLMs within multilingual and multicultural contexts \cite{ahuja-etal-2023-mega,ahuja-etal-2024-megaverse,faisal-etal-2024-dialectbench,watts-etal-2024-pariksha,chiu-etal-2025-culturalbench}. Traditionally, these benchmarks were developed by translating English versions into various languages. However, due to the loss of linguistic and cultural context during translation, new benchmarks specific to different languages and cultures are now being created. However, such benchmarks are few in number, and several of the older ones are contaminated in training data \cite{ahuja-etal-2024-megaverse,oren2024proving,deng-etal-2024-investigating}. Thus, there is a need for new benchmarks that can test the abilities of models in real-world multilingual settings.

LLMs are employed in various fields, including critical areas like healthcare. \citet{jin2024better} translate an English healthcare dataset into Spanish, Chinese, and Hindi, and demonstrate that performance declines in these languages compared to English. This highlights the necessity of examining LLMs more thoroughly in multilingual contexts for these important uses.

In this study, we conduct the first comprehensive assessment of multilingual models within a real-world healthcare context. We evaluate responses from 24 multilingual and Indic models using 750 questions posed by users of a health chatbot in five languages (Indian-English and four Indic languages). All the models being evaluated function within the same Retrieval Augmented Generation (RAG) framework \cite{NEURIPS2020_6b493230,karpukhin-etal-2020-dense}, and their outputs are compared to doctor-verified ground truth responses. We evaluate LLM responses on four metrics curated for our application, including factual correctness, semantic similarity, coherence, and conciseness, and present leaderboards for each metric, as well as an overall leaderboard. We use human evaluation and automated methods (LLMs-as-a-judge) to compute these metrics by comparing LLM responses with ground-truth reference responses or assessing the responses in a reference-free manner.

Our results suggest that models vary significantly in their performance, with some smaller models outperforming larger ones. Factual Correctness is generally lower for non-English queries compared to English queries. We observe that instruction-tuned Indic models do not always perform well on Indic language queries. Our dataset contains several instances of code-mixed and culturally-relevant queries, which models sometimes struggle to answer. The contributions of our work are as follows:

\begin{itemize}
    \item We evaluate 24 models (proprietary as well as open weights) in a healthcare setting using queries provided by patients using a medical chatbot. This guarantees that our dataset is not contaminated in the training data of any of the models we evaluate.
    \item We curate a dataset of queries from multilingual users that spans multiple languages. The queries feature language typical of multilingual communities, such as code-switching, which is seldom in translated datasets, making ours a more realistic dataset for model evaluation.
    \item We evaluate several models in an identical RAG setting, making it possible to compare models fairly. The RAG setting is a popular configuration that numerous models are being deployed in for real-world applications.
    \item We establish relevant metrics for our application and determine an overall combined metric by consulting doctors working on the medical chatbot project.
    \item We perform assessments (with and without ground truth references) using LLM-as-a-judge and conduct human evaluations on a subset of the models and data to confirm the validity of the LLM assessment.
\end{itemize}
\section{Related Works}
\label{sec:related_works}

\paragraph{Healthcare Chatbots in India} Within the Indian context, the literature has documented great diversity in health-seeking and health communication behaviors based on gender \cite{Das_Angeli_Krumeich_Schayck_2018}, varying educational status, poor functional literacy, cultural context \cite{Islary_2018}, stigmas \cite{info:doi/10.2196/29969}, etc. This diversity in behavior may translate to people’s use of medical chatbots, which are increasingly reaching hundreds of Indian patients at the margins of the healthcare system \cite{mishra-etal-2023-hindi}. These bots solicit personal health information directly from patients in their native Indic languages or in Indic English. For example, \citet{Ramjee_2024} find that their \textsc{CataractBot} deployed in Bangalore, India, yields patient questions on topics such as surgery, preoperative preparation, diet, exercise, discharge, medication, pain management, etc. \citet{info:doi/10.2196/29969} find that Indian people share ``\textit{deeply personal questions and concerns about sexual and reproductive health}'' with their chatbot SnehAI. \citet{Yadav_2019} find that queries to chatbots are ``\textit{embedded deeply into a community's myths and existing belief systems}'' while \cite{Xiao_2023} note that patients have difficulties finding health information at an appropriate level for them to comprehend. Hence, LLMs powering medical chatbots in India and other low and middle-income countries are challenged to respond lucidly to medical questions that are asked in ways that may be hyperlocal to the patient context. Few works have documented how LLMs react to this linguistic diversity in the medical domain. Our work aims to bridge this gap.

\paragraph{Multilingual and RAG evaluation} Several previous studies have conducted in-depth evaluation of Multilingual capabilities of LLMs by evaluating across standard tasks \cite{srivastava2022beyond,liang2023holistic,ahuja-etal-2023-mega,ahuja-etal-2024-megaverse,asai-etal-2024-buffet,lai-etal-2023-chatgpt,robinson-etal-2023-chatgpt}, with a common finding that current LLMs only have a limited multilingual capacity \cite{ochieng-etal-2025-beyond}. Other works \cite{watts-etal-2024-pariksha,leong2023bhasa} include evaluating LLMs on creative and generative tasks. \citet{10.1145/3626772.3657957} state that evaluating RAG models requires a joint evaluation of the retrieval and generated output. Recent works such as \citet{Chen_Lin_Han_Sun_2024,chirkova-etal-2024-retrieval} benchmark LLMs as RAG  models in bilingual and multilingual setups. Lastly, several tools and benchmarks have also been built for automatic evaluation of RAG, even in medical domains \cite{es-etal-2024-ragas,tang2024multihoprag,xiong-etal-2024-benchmarking,xiong2024improving}, and we refer the readers to \citet{yu2024evaluation} for such a comprehensive list and survey.

\paragraph{LLM-based Evaluators} With the advent of large-scale instruction following capabilities in LLMs, automatic evaluations with the help of these models is being preferred \cite{pombal2025mprometheus,kim2024prometheus,kim-etal-2024-prometheus,doddapaneni-etal-2025-cross,liu2024omgeval,shen-etal-2023-large,kocmi-federmann-2023-large}. However, it has been shown that it is optimal to assess these evaluations in tandem with human annotations as LLMs can provide inflated scores \cite{hada-etal-2024-large,hada-etal-2024-metal,watts-etal-2024-pariksha}. Other works \cite{zheng2023judging,watts-etal-2024-pariksha} have employed GPT-4 alongside human evaluators on leaderboards to assess other LLMs. \citet{ning2025pico} proposed an innovative approach using LLMs for peer review, where models evaluate each other's outputs. However, a study by \citet{doddapaneni-etal-2024-finding} highlighted the limitations of LLM-based evaluators, revealing their inability to reliably detect subtle drops in input quality during evaluations, raising concerns about their precision and dependability for fine-grained assessments. In this work, we use LLM-based evaluators both with and without ground-truth references, and also use human evaluation to validate LLM-based evaluation.
\section{Methodology}
\label{sec:methodology}

In this study, we leveraged a dataset collected from a deployed medical chatbot. Here, we provide an overview of the question dataset, the knowledge base employed for answering those questions, the process for generating responses, and the evaluation framework.

\paragraph{Data} \label{para:data} The real-world test data was collected by our collaborators as part of an ongoing research effort that designed and deployed a medical chatbot, hereafter referred to as \textsc{HealthBot}, to patients scheduled for cataract surgery at a large hospital in urban India. An Ethics approval was obtained from our institution before conducting this work, and once enrolled in the study and consent was obtained, both the patient and their accompanying family member or attendant were instructed on how to use \textsc{HealthBot} on WhatsApp. Through this instructional phase, they were informed that questions could be asked by voice or by text, in one of 5 languages - English, Hindi, Kannada, Tamil, Telugu. The workflow of chatting with \textsc{HealthBot} was as follows: Patients sent questions through the WhatsApp interface to \textsc{HealthBot}. Their questions were transcribed automatically and later translated using an off-the-shelf translator \cite{gala2023indictrans,gumma-etal-2025-towards, nllbteam2022languageleftbehindscaling} into English if needed, after which GPT-4 was used to produce an initial response by performing RAG on the documents in the knowledge base (KB). This initial response was passed to doctors who reviewed, validated, and, if needed, edited the answer. The doctor-approved answer is referred to as the ground truth (GT) response associated with the patient query. 

Our evaluation dataset was curated from this data by including all questions sent to \textsc{HealthBot} along with their associated GT response. Exclusion criteria removed exact duplicate questions, those with personally identifying information, and those not relevant to health. Additionally, for this work, we only consider questions to which the GPT-4 answer was directly approved by the expert as the ``\textit{correct and complete answer}" without additional editing on the doctors' part. The final dataset contained 749 questions and GT answer pairs that were sent to \textsc{HealthBot} between December 2023 to June 2024. In the pool, 666 questions were in English, 19 in Hindi, 27 in Tamil, 14 in Telugu, and 23 in Kannada. Note that queries written in the script of a specific language were classified as belonging to that language. For code-mixed and Romanized queries, we determined whether they were English or non-English based on the matrix language of the query. 

The evaluation dataset consists of queries that (1) have misspelled English words, (2) are code-mixed, (3) represent non-native English, (4) are relevant to the patient’s cultural context, and (5) are specific to the patient’s condition. We provide some examples of each of these categories. 

Examples of misspelled queries include questions such as ``\textit{How long should saving not be done after surgery?}'' where the patient intended to ask about shaving, and ``\textit{Sarjere is don mam?}'' which the attendant used to inquire about the patient's discharge status. Instances of code mixing can be seen in phrases like ``\textit{Agar operation ke baad pain ho raha hai, to kya karna hai?}'' meaning ``\textit{If there is pain after the surgery, what should I do?}'' in Hindi-English. Other examples include ``\textit{Can I eat before the kanna operation?}'' where ``\textit{kanna}'' means eye in Tamil, and ``\textit{kanna operation}'' is a well-understood, common way of referring to cataract surgery, and ``\textit{In how many days can a patient take Karwat?}'' where ``\textit{Karwat}'' means ``turning over in sleep" in Hindi. 

Indian English was used in a majority of the English queries, making the phrasing of questions different from what they would be with native English speech. Examples are as follows - ``\textit{Because I have diabetes sugar problem I am worried much}'', ``\textit{Why to eat light meal only? What comes under light meal?}'' and ``\textit{Is the patient should be in dark room after surgery?}'' Taking a shower was commonly referred to as ``\textit{taking a bath}'', and eye glasses were commonly referred to as ``\textit{goggles}'', ``\textit{spex}'' or ``\textit{spectacles}''. 

Culturally-relevant questions were also many in number, for example, questions about specific foods were asked like ``\textit{Can he take chapati, Puri etc on the day of surgery?}'' and ``\textit{Can I eat non veg after surgery?}'' (``\textit{non-veg}'' is a term used in Indian English to denote eating meat). Questions about yoga were asked, like ``\textit{How long after the surgery should the Valsalva maneuver be avoided?}'' and ``\textit{Are there any specific yoga poses I can do?}''. The notion of a patient’s native place or village was brought up in queries such as ``\textit{If a person gets operated here and then goes to his native place and if some problem occurs what shall he do ?}'' or ``\textit{Can she travel by car with AC for 100 kms ?}''.


\paragraph{Knowledge Base} The documents populating the knowledge base (KB) were initially curated by doctors at the hospital where \textsc{HealthBot} was deployed. This consisted of 12 PDF documents that were converted into text files and manually error checked. The documents included Standard Operating Procedure manuals, standard treatment guidelines, consent forms, frequently-asked-question documents, insurance information, etc. Following this initial curation, doctors who were with \textsc{HealthBot} were able to select question-answer pairs to be added to the KB after the bot was deployed. In this manner, the knowledge available to GPT-4 in the KB grew over time. Therefore, every question that was asked by patients was associated with a different version of the KB being used for answer generation. This detail was incorporated into our evaluation in order to compare the verified ground truth data with the generated response in an accurate manner. All KB documents were chunked to a maximum length of 1000 tokens, and embedded in a VectorDB using the \textsc{Text-Embedding-Ada-002}. Subsequently, for each query, the top 3 most relevant chunks are extracted, and the models are queried with this data. 

\paragraph{Models} We chose 24 models, including proprietary multilingual models, as well as Open-weights multilingual and Indic language models for our evaluation. A full list of models can be found in Table \ref{tab: model_list}.
\begin{table*}[h]
\centering 
\small
\begin{tabular}{@{}lcc@{}}
\toprule
\textbf{Models} & \textbf{\begin{tabular}[c]{@{}c@{}}Languages \\ Tested\end{tabular}} & \textbf{Availability} \\ \midrule
\textsc{GPT-4} & All & Proprietary \\
\textsc{GPT-4o} & All & Proprietary \\
\textsc{microsoft/Phi-3.5-MoE-instruct} & All & Open-weights \\
\textsc{CohereForAI/c4ai-command-r-plus-08-2024} & All & Open-weights \\
\textsc{Qwen/Qwen2.5-72B-Instruct} & All & Open-weights \\
\textsc{CohereForAI/aya-23-35B} & All & Open-weights \\
\textsc{mistralai/Mistral-Large-Instruct-2407} & All & Open-weights \\
\textsc{google/gemma-2-27b-it} & All & Open-weights \\
\textsc{meta-llama/Meta-Llama-3.1-70B-Instruct} & All & Open-weights \\
\textsc{GenVRadmin/llama38bGenZ\_Vikas-Merged} & All & Indic \\
\textsc{GenVRadmin/AryaBhatta-GemmaOrca-Merged} & All & Indic \\
\textsc{GenVRadmin/AryaBhatta-GemmaUltra-Merged} & All & Indic \\
\textsc{GenVRadmin/AryaBhatta-GemmaGenZ-Vikas-Merged} & All & Indic \\
\textsc{Telugu-LLM-Labs/Indic-gemma-7b-finetuned-sft-Navarasa-2.0} & All & Indic \\
\textsc{ai4bharat/Airavata} & En, Hi & Indic \\
\textsc{Cognitive-Lab/LLama3-Gaja-Hindi-8B-v0.1} & En, Hi & Indic \\
\textsc{BhabhaAI/Gajendra-v0.1} & En, Hi & Indic \\
\textsc{manishiitg/open-aditi-hi-v4} & En, Hi & Indic \\
\textsc{abhinand/tamil-llama-7b-instruct-v0.2} & En, Ta & Indic \\
\textsc{abhinand/telugu-llama-7b-instruct-v0.1} & En, Te & Indic \\
\textsc{Telugu-LLM-Labs/Telugu-Llama2-7B-v0-Instruct} & En, Te & Indic \\
\textsc{Tensoic/Kan-Llama-7B-SFT-v0.5} & En, Ka & Indic \\
\textsc{Cognitive-Lab/Ambari-7B-Instruct-v0.2} & En, Ka & Indic \\
\textsc{GenVRadmin/Llamavaad} & En, Hi & Indic \\ \bottomrule
\end{tabular}
\caption{List of models tested. En = English, Hi = Hindi, Ka = Kannada, Ta = Tamil, Te = Telugu, and ``All" refers to all the aforementioned languages. All Indic models are open-weight as well, but are predominantly fine-tuned with open-source Indic data. We can only hypothesize that most of the proprietary and open-weight models mentioned above also have some fraction of Indic data in their training data, but no official information about the language mixture is released.}
\label{tab: model_list}
\end{table*}

\paragraph{Response Generation} We use the standard RAG strategy to elicit responses from all the models. Each model is asked to respond to the given query by extracting the appropriate pieces of text from the knowledge-base chunks. During prompting, we segregate the chunks into \textsc{RawChunks} and \textsc{KBUpdateChunks} symbolizing the data from the standard sources, and the KB updates. Then the model is explicitly instructed to prioritize the information from the most recent sources, i.e., the \textsc{KBUpdateChunks} (if they are available). The exact prompt used for generation is provided in Appendix \ref{sec:prompts}. Note that each model gets the same \textsc{RawChunks} and \textsc{KBUpdateChunks}, which are also the same that are given to the GPT-4 model in the \textsc{HealthBot}, based on which the GT responses are verified.

\paragraph{Response Evaluation} We used both human and automated evaluation to evaluate the performance of models in the setup described above. GPT-4o was employed as an LLM evaluator. We prompted the model separately to judge each metric, as \citet{hada-etal-2024-large,hada-etal-2024-metal} show that individual calls reduce interaction and influence among them and their evaluations.

\subparagraph{LLM-based Evaluation} In consultation with domain experts working on the \textsc{HealthBot}, we curated metrics that are relevant for our application. We limit ourselves to 3 classes \textbf{(Good - 2, Medium - 1, Bad - 0)} for each metric, as a larger number of classes could hurt interpretability and lower LLM-evaluator performance. The prompts used for each of our metrics are available in Appendix \ref{sec:prompts}, and a general overview is provided below. 

\begin{itemize}
\item \factualcorrectness\ (FC): As \citet{doddapaneni-etal-2024-finding} had shown that LLM-based evaluators fail to identify subtle factual inaccuracies, we curate a separate metric to double-check facts like dates, numbers, procedures, and medicine names. 
\item \semanticsimilarity\ (SS): Similarly, we formulate another metric to specifically analyse if both the prediction and the ground-truth response convey the same information semantically, especially when they are in different languages.
\item \coherence\ (COH): This metric evaluates if the model was able to stitch together appropriate pieces of information from the three data chunks provided to yield a coherent response. 
\item \conciseness\ (CON): Since the knowledge base chunks extracted and provided to the model can be quite large, with important facts embedded at different positions, we build this metric to assess the ability of the model to extract and compress all these bits of information relevant to the query into a crisp response.
\end{itemize}

Among the metrics presented above, \factualcorrectness\ and \semanticsimilarity\ use the GT response verified by doctors as a reference, while \coherence\ and \conciseness\ are reference-free metrics. To arrive at a combined score for each model, we asked two doctors who collaborate on the \textsc{HealthBot} to assign weights to the first four metrics according to their importance and used an average of the percentages for each metric as the final coefficient to compute the \aggregate\ (AGG). Both doctors gave the maximum weight to \factualcorrectness\ followed by \semanticsimilarity\, while \coherence\ and \conciseness\ were given lower and equal weightage.

\subparagraph{Human Evaluation} Following previous works \cite{hada-etal-2024-large,hada-etal-2024-metal,watts-etal-2024-pariksha}, we augment the LLM evaluation with human evaluation and draw correlations between the LLM evaluator and human evaluation for a subset of the models (\textsc{Phi-3.5-MoE-instruct}, \textsc{Mistral-Large-Instruct-2407}, \textsc{gpt-4o}, \textsc{Meta-Llama-3.1-70B-Instruct}, \textsc{Indic-gemma-7b-finetuned-sft-Navarasa-2.0}). These models were selected based on results from early automated evaluations, covering a range of scores and representing models of interest. The human annotators were employed by \redactedname, a data annotation company, and were all native speakers of Indian languages that we evaluated. We selected a sample of 100 queries from English and all the queries from Indic languages for annotation, yielding a total of 183 queries. Each instance was annotated by one annotator for \semanticsimilarity\ between the model's response and the GT response provided by the doctor. The annotations began with a briefing about the task, and each of them was given a sample test task and was provided with some guidance based on their difficulties and mistakes. Finally, the annotators were asked to evaluate the model response based on the metric\footnote{The formulation and wording of the metric were slightly simplified for the annotators to better understand it.}, query, and ground-truth response on a scale of 0 to 2, similar to the LLM-evaluator.
\section{Results}
\label{sec:results}

In this section, we present the outcomes of both the LLM and human evaluations. We begin by examining the average scores across all our metrics, including the combined metric for English queries, followed by results for queries in other languages. Next, we examine the ranking of models based on scores given by human annotators and compare these rankings based on scores provided by the LLM evaluator. Lastly, we conduct a qualitative analysis of the outcomes and describe noteworthy findings.

\paragraph{LLM evaluator results} We see from Table \ref{tab:en-results} that for English, the best performing model is the \textsc{Qwen2.5-72B-Instruct} model across all metrics. Note that it is expected that GPT-4 performs well, as the ground truth responses are based on responses generated by \textsc{GPT-4}. The \textsc{Phi-3.5-MoE-instruct} model also performs well on all metrics, followed by \textsc{Mistral-Large-Instruct-2407} and \textsc{open-aditi-hi-v4}, which is the only Indic model that performs near the top even for English queries. Surprisingly, the \textsc{Meta-Llama-3.1-70B-Instruct} model performs worse than expected on this task, frequently regurgitating the entire prompt that was provided. In general, all models get higher scores on conciseness, and many models do well on coherence. 

\begin{table*}[h]
\centering 
\small
\begin{tabular}{@{}lc|cccc@{}}
\toprule
\textbf{Model} & \multicolumn{1}{c}{\textbf{AGG}} & \multicolumn{1}{c}{\textbf{COH}} & \multicolumn{1}{c}{\textbf{CON}} & \multicolumn{1}{c}{\textbf{FC}} & \multicolumn{1}{c}{\textbf{SS}} \\ \midrule
\textsc{Qwen2.5-72B-Instruct} & 1.46 & 1.86 & 1.96 & 1.62 & 1.43 \\
\textsc{gpt-4} & 1.40 & 1.71 & 1.95 & 1.56 & 1.36 \\
\textsc{Phi-3.5-MoE-instruct} & 1.29 & 1.65 & 1.93 & 1.43 & 1.22 \\
\textsc{Mistral-Large-Instruct-2407} & 1.29 & 1.60 & 1.95 & 1.42 & 1.24 \\
\textsc{open-aditi-hi-v4} & 1.27 & 1.69 & 1.85 & 1.37 & 1.22 \\
\textsc{Llamavaad} & 1.16 & 1.34 & 0.97 & 1.36 & 1.20 \\
\textsc{AryaBhatta-GemmaGenZ-Vikas-Merged} & 1.12 & 1.48 & 1.65 & 1.22 & 1.07 \\
\textsc{Kan-Llama-7B-SFT-v0.5} & 1.01 & 1.39 & 1.64 & 1.07 & 0.97 \\
\textsc{gemma-2-27b-it} & 1.00 & 1.28 & 1.88 & 1.07 & 0.91 \\
\textsc{AryaBhatta-GemmaOrca-Merged} & 0.97 & 1.32 & 1.62 & 1.03 & 0.92 \\
\textsc{LLama3-Gaja-Hindi-8B-v0.1} & 0.91 & 0.63 & 1.65 & 1.09 & 0.98 \\
\textsc{gpt-4o} & 0.91 & 1.08 & 1.78 & 0.98 & 0.87 \\
\textsc{aya-23-35B} & 0.91 & 1.09 & 1.65 & 1.00 & 0.83 \\
\textsc{Gajendra-v0.1} & 0.88 & 1.21 & 1.38 & 0.93 & 0.85 \\
\textsc{c4ai-command-r-plus-08-2024} & 0.82 & 1.15 & 1.48 & 0.85 & 0.74 \\
\textsc{tamil-llama-7b-instruct-v0.2} & 0.81 & 1.13 & 1.50 & 0.83 & 0.75 \\
\textsc{Airavata} & 0.80 & 1.03 & 1.38 & 0.85 & 0.78 \\
\textsc{Ambari-7B-Instruct-v0.2} & 0.73 & 0.86 & 1.11 & 0.76 & 0.82 \\
\textsc{Meta-Llama-3.1-70B-Instruct} & 0.65 & 0.55 & 1.12 & 0.77 & 0.67 \\
\textsc{Telugu-Llama2-7B-v0-Instruct} & 0.51 & 0.60 & 1.12 & 0.53 & 0.53 \\
\textsc{llama38bGenZ\_Vikas-Merged} & 0.51 & 0.52 & 1.09 & 0.55 & 0.53 \\
\textsc{Indic-gemma-7b-finetuned-sft-Navarasa-2.0} & 0.35 & 0.32 & 0.53 & 0.40 & 0.39 \\
\textsc{AryaBhatta-GemmaUltra-Merged} & 0.32 & 0.38 & 1.19 & 0.31 & 0.27 \\
\textsc{telugu-llama-7b-instruct-v0.1} & 0.04 & 0.00 & 0.58 & 0.03 & 0.00 \\ \bottomrule
\end{tabular}
\caption{Metric-wise scores for English.}
\label{tab:en-results}
\end{table*}

\begin{table*}[h]
\small
\centering 
\begin{tabular}{@{}lc|cccc@{}}
\toprule
\textbf{Model} & \textbf{AGG} & \textbf{COH} & \textbf{CON} & \textbf{FC} & \textbf{SS} \\ \midrule
\textsc{gpt-4} & 1.21 & 1.74 & 1.79 & 1.26 & 1.16 \\
\textsc{Qwen2.5-72B-Instruct} & 1.20 & 1.89 & 1.95 & 1.21 & 1.11 \\
\textsc{Mistral-Large-Instruct-2407} & 1.18 & 1.53 & 1.79 & 1.26 & 1.16 \\
\textsc{gemma-2-27b-it} & 0.93 & 1.11 & 1.89 & 1.05 & 0.79 \\
\textsc{aya-23-35B} & 0.92 & 0.95 & 1.79 & 1.05 & 0.84 \\
\textsc{AryaBhatta-GemmaGenZ-Vikas-Merged} & 0.89 & 1.11 & 1.32 & 1.00 & 0.84 \\
\textsc{Phi-3.5-MoE-instruct} & 0.81 & 1.11 & 1.74 & 0.79 & 0.79 \\
\textsc{gpt-4o} & 0.76 & 0.74 & 1.79 & 0.84 & 0.74 \\
\textsc{AryaBhatta-GemmaOrca-Merged} & 0.64 & 1.00 & 1.21 & 0.58 & 0.68 \\
\textsc{Airavata} & 0.63 & 0.84 & 1.26 & 0.68 & 0.53 \\
\textsc{LLama3-Gaja-Hindi-8B-v0.1} & 0.60 & 0.79 & 1.26 & 0.63 & 0.53 \\
\textsc{open-aditi-hi-v4} & 0.56 & 0.89 & 1.00 & 0.47 & 0.63 \\
\textsc{Llamavaad} & 0.55 & 0.47 & 0.21 & 0.68 & 0.63 \\
\textsc{c4ai-command-r-plus-08-2024} & 0.52 & 0.95 & 1.47 & 0.47 & 0.37 \\
\textsc{Meta-Llama-3.1-70B-Instruct} & 0.48 & 0.47 & 1.16 & 0.53 & 0.47 \\
\textsc{Gajendra-v0.1} & 0.38 & 0.47 & 0.68 & 0.37 & 0.42 \\
\textsc{llama38bGenZ\_Vikas-Merged} & 0.32 & 0.21 & 1.00 & 0.32 & 0.37 \\
\textsc{AryaBhatta-GemmaUltra-Merged} & 0.31 & 0.37 & 1.00 & 0.32 & 0.26 \\
\textsc{Indic-gemma-7b-finetuned-sft-Navarasa-2.0} & 0.24 & 0.11 & 0.53 & 0.26 & 0.32 \\ \bottomrule
\end{tabular}
\caption{Metric-wise scores for Hindi}
\label{tab:hin-scores}
\end{table*}
\begin{table*}[h]
\small
\centering
\begin{tabular}{@{}lc|cccc@{}}
\toprule
\textbf{Model} & \textbf{AGG} & \textbf{COH} & \textbf{CON} & \textbf{FC} & \textbf{SS} \\ \midrule
\textsc{Qwen2.5-72B-Instruct} & 1.29 & 1.87 & 1.96 & 1.35 & 1.22 \\
\textsc{gpt-4} & 1.18 & 1.78 & 1.96 & 1.30 & 0.91 \\
\textsc{Mistral-Large-Instruct-2407} & 1.09 & 1.39 & 1.96 & 1.22 & 0.96 \\
\textsc{gemma-2-27b-it} & 0.92 & 1.30 & 1.91 & 1.04 & 0.65 \\
\textsc{gpt-4o} & 0.88 & 0.96 & 2.00 & 1.00 & 0.74 \\
\textsc{AryaBhatta-GemmaOrca-Merged} & 0.51 & 0.57 & 1.13 & 0.52 & 0.52 \\
\textsc{Meta-Llama-3.1-70B-Instruct} & 0.48 & 0.43 & 0.78 & 0.57 & 0.48 \\
\textsc{Kan-Llama-7B-SFT-v0.5} & 0.47 & 0.52 & 1.04 & 0.48 & 0.48 \\
\textsc{llama38bGenZ\_Vikas-Merged} & 0.47 & 0.52 & 1.00 & 0.43 & 0.57 \\
\textsc{Indic-gemma-7b-finetuned-sft-Navarasa-2.0} & 0.24 & 0.35 & 0.39 & 0.26 & 0.22 \\
\textsc{Phi-3.5-MoE-instruct} & 0.20 & 0.26 & 1.22 & 0.17 & 0.09 \\
\textsc{AryaBhatta-GemmaUltra-Merged} & 0.13 & 0.17 & 0.70 & 0.09 & 0.13 \\
\textsc{Ambari-7B-Instruct-v0.2} & 0.05 & 0.04 & 0.13 & 0.04 & 0.09 \\ \bottomrule
\end{tabular}
\caption{Metric-wise scores for Kannada}
\label{tab:kan-scores}
\end{table*}

\begin{table*}[h]
\small
\centering 
\begin{tabular}{@{}lc|cccc@{}}
\toprule
\textbf{Model} & \textbf{AGG} & \textbf{COH} & \textbf{CON} & \textbf{FC} & \textbf{SS} \\ \midrule
\textsc{Qwen2.5-72B-Instruct} & 1.29 & 1.87 & 1.96 & 1.35 & 1.22 \\
\textsc{gpt-4} & 1.18 & 1.78 & 1.96 & 1.30 & 0.91 \\
\textsc{Mistral-Large-Instruct-2407} & 1.09 & 1.39 & 1.96 & 1.22 & 0.96 \\
\textsc{gemma-2-27b-it} & 0.92 & 1.30 & 1.91 & 1.04 & 0.65 \\
\textsc{gpt-4o} & 0.88 & 0.96 & 2.00 & 1.00 & 0.74 \\
\textsc{AryaBhatta-GemmaOrca-Merged} & 0.51 & 0.57 & 1.13 & 0.52 & 0.52 \\
\textsc{Meta-Llama-3.1-70B-Instruct} & 0.48 & 0.43 & 0.78 & 0.57 & 0.48 \\
\textsc{Kan-Llama-7B-SFT-v0.5} & 0.47 & 0.52 & 1.04 & 0.48 & 0.48 \\
\textsc{llama38bGenZ\_Vikas-Merged} & 0.47 & 0.52 & 1.00 & 0.43 & 0.57 \\
\textsc{Indic-gemma-7b-finetuned-sft-Navarasa-2.0} & 0.24 & 0.35 & 0.39 & 0.26 & 0.22 \\
\textsc{Phi-3.5-MoE-instruct} & 0.20 & 0.26 & 1.22 & 0.17 & 0.09 \\
\textsc{AryaBhatta-GemmaUltra-Merged} & 0.13 & 0.17 & 0.70 & 0.09 & 0.13 \\
\textsc{Ambari-7B-Instruct-v0.2} & 0.05 & 0.04 & 0.13 & 0.04 & 0.09 \\ \bottomrule
\end{tabular}
\caption{Metric-wise scores for Tamil}
\label{tab:tam-scores}
\end{table*}
\begin{table*}[h]
\small
\centering 
\begin{tabular}{@{}lc|cccc@{}}
\toprule
\textbf{Model} & \textbf{AGG} & \textbf{COH} & \textbf{CON} & \textbf{FC} & \textbf{SS} \\ \midrule
\textsc{gpt-4} & 1.14 & 1.64 & 2.00 & 1.29 & 0.86 \\
\textsc{Qwen2.5-72B-Instruct} & 1.11 & 1.57 & 1.71 & 1.29 & 0.86 \\
\textsc{Mistral-Large-Instruct-2407} & 1.03 & 1.36 & 2.00 & 1.14 & 0.86 \\
\textsc{gemma-2-27b-it} & 0.91 & 1.21 & 2.00 & 1.00 & 0.71 \\
\textsc{Meta-Llama-3.1-70B-Instruct} & 0.61 & 0.43 & 1.00 & 0.79 & 0.57 \\
\textsc{gpt-4o} & 0.54 & 0.57 & 1.86 & 0.57 & 0.43 \\
\textsc{Phi-3.5-MoE-instruct} & 0.44 & 0.57 & 1.86 & 0.43 & 0.29 \\
\textsc{llama38bGenZ\_Vikas-Merged} & 0.33 & 0.14 & 1.50 & 0.36 & 0.29 \\
\textsc{AryaBhatta-GemmaOrca-Merged} & 0.29 & 0.29 & 0.93 & 0.29 & 0.29 \\
\textsc{AryaBhatta-GemmaUltra-Merged} & 0.26 & 0.29 & 1.71 & 0.21 & 0.14 \\
\textsc{Indic-gemma-7b-finetuned-sft-Navarasa-2.0} & 0.19 & 0.29 & 0.57 & 0.21 & 0.07 \\
\textsc{telugu-llama-7b-instruct-v0.1} & 0.09 & 0.00 & 1.71 & 0.00 & 0.00 \\
\textsc{Telugu-Llama2-7B-v0-Instruct} & 0.00 & 0.00 & 0.00 & 0.00 & 0.00 \\ \bottomrule
\end{tabular}
\caption{Metric-wise scores for Telugu}
\label{tab:tel-scores}
\end{table*}

For the non-English queries, which are far fewer in number compared to English (Tables \ref{tab:hin-scores}, \ref{tab:tam-scores}, \ref{tab:tel-scores}, \ref{tab:kan-scores}), we find that models such as \textsc{Aya-23-35B} perform near the top for Hindi along with proprietary and large open weights models such as \textsc{Qwen2.5-72B-Instruct} and \textsc{Mistral-Large-Instruct-2407}, outperforming many of the fine-tuned Indic LLMs. The \textsc{gemma-2-27b-it} model also outperforms many Indic models in the Indic setting, compared to its performance in English. This shows that some instruction-tuned Indic LLMs may not perform well in the RAG setting. We also find that compared to English, models get lower values on FC on Indic queries, which is concerning, as it is rated as the most important metric by doctors.

\begin{figure}[h]
    \centering
    \begin{subfigure}[b]{\linewidth}
        \centering
        \includegraphics[width=\linewidth]{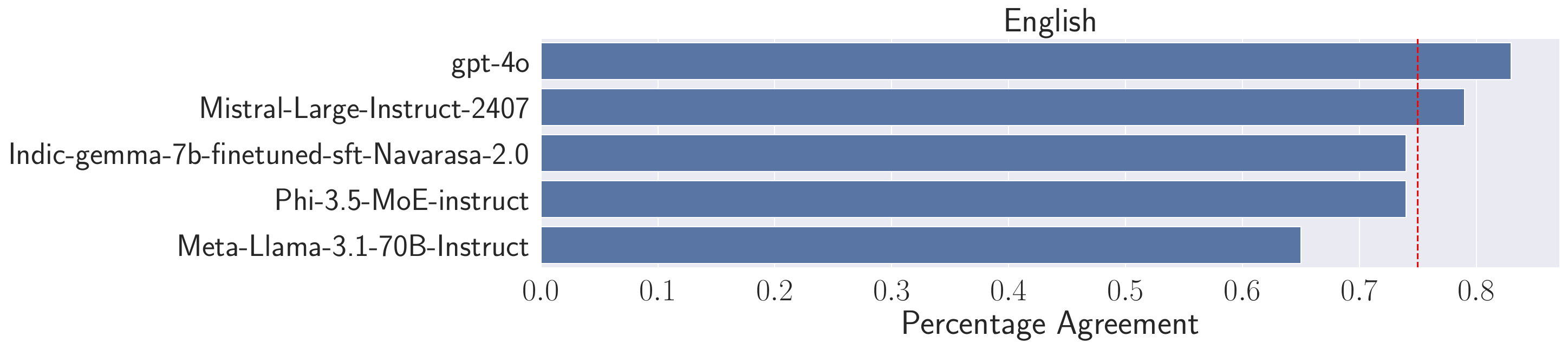}
        \caption{English.}
        \label{fig:PA-en}
    \end{subfigure}
    \hfill
    \begin{subfigure}[b]{\linewidth}
        \centering
        \includegraphics[width=\linewidth]{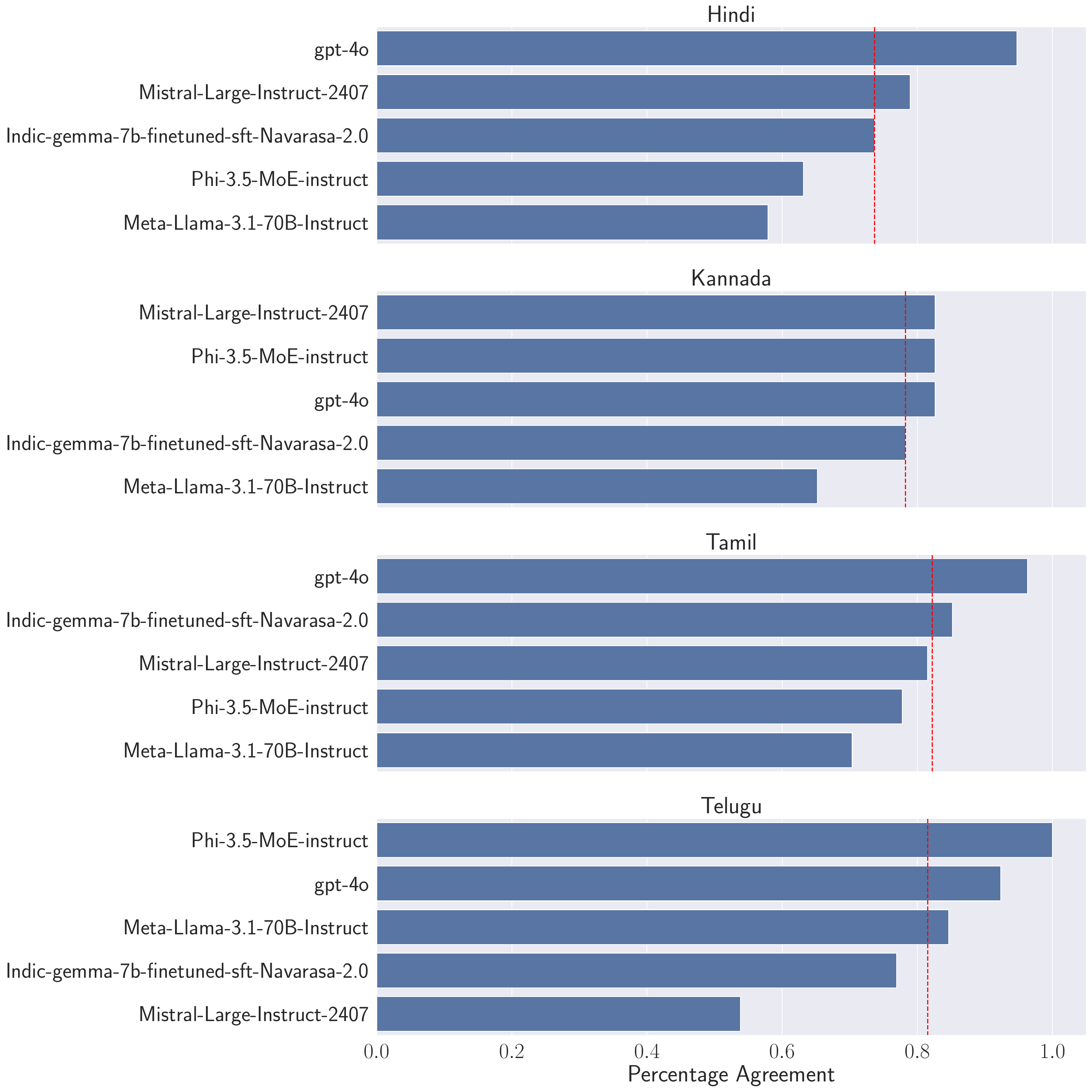}
        \caption{Indic languages}
        \label{fig:pa-indic}
    \end{subfigure}
    \caption{Percentage Agreement (PA) between human and LLM-evaluators. The red line indicates the average PA across models.}
    \label{fig:pa-combined}
\end{figure}

\paragraph{Comparison of human and LLM evaluators}
We perform human evaluation on five models on the \semanticsimilarity\ (SS) task and compare human and LLM evaluation by inspecting the ranking of the models in Appendix \ref{sec:human_llm_comparison}. We find that for all languages except Telugu, we get identical rankings of all models. Additionally, we also measure the Percentage Agreement (PA) between the human and LLM-evaluator, details of which can be found in the Figure \ref{fig:pa-combined} and find it to be consistently higher than 0.7 on average across all languages and models. This shows the reliability of our LLM-based evaluation for \semanticsimilarity\, which uses the GT response as a reference.

\paragraph{Qualitative Analysis} One of the authors of the paper performed a qualitative analysis of responses from the evaluated LLMs on 100 selected patient questions. The questions were chosen to cover a range of medical topics and languages. Thematic analysis involved (1) initial familiarization with the queries and associated LLM responses, (2) theme identification, where 5 themes were generated, and (3) thematic coding, where the generated themes were applied to the 100 question-answer pairs. We briefly summarize these results:

The five generated themes across queries were (1) misspelling of English words, (2) code-mixing, (3) non-native English, (4) relevance to cultural context, and (5) specificity to the patient’s condition.

For queries that involve misspellings (such as ``\textit{saving}'' and ``\textit{sarjere}'' mentioned in Paragraph \ref{para:data}), many evaluated LLMs were not able to come up with an appropriate response. For the query with the word ``\textit{saving}", responses varied from ``\textit{The patient should not be saved for more than 15 days after the surgery}'' to ``\textit{Saving should not be done after surgery}'' to ``\textit{You should not strain to pass motion for 15 days after the surgery. If you are constipated, it is recommended to consult the doctor}''. All of these responses deviate from the \textsc{GPT-4} generated GT, which said ``\textit{You can have a shave after the cataract surgery. However, you should avoid having a head bath or shampoo for 15 days post-surgery.}''

In cases of code mixing and Indian English, LLMs were more robust in their responses than to misspellings. The term ``\textit{Kanna operation}'' was well understood by most models, and Hinglish queries were responded to in a manner that mirrored the GT answer. However, less common terms were not understood when mixed with English. The query with the word ``\textit{Karwat}'' as mentioned in Paragraph \ref{para:data} received responses ranging from ``\textit{you can start cooking after 1 week}'' to ``\textit{I'm sorry, but I cannot provide an answer to your question. The information you are seeking is not relevant to cataract surgery or the guidelines for post-operative care}'' to ``\textit{be careful when children get near you}''. Most of the evaluated LLMs understood the use of ``\textit{sugar}'' in reference to diabetes, as well as sentences following different syntax than would be common in native English.   

The responses for culturally-relevant questions varied greatly between evaluated LLMs. For example, to the question on appropriateness of chapati and puri on the day of surgery, some LLMs approved, saying ``\textit{Yes, he can take chapati, Puri etc on the day of cataract surgery}'' while others were against this, saying ``\textit{You should have a light meal before the surgery. Avoid heavy or oily foods like chapati and Puri on the day of your cataract surgery. It's best to stick to easily digestible foods. If you have any specific dietary concerns, please discuss them with your healthcare team}''. Questions relating to returning to a ``\textit{native place}'' were met with refusals by around half of the evaluated LLMs. 

Questions that were specific to the patient's condition were also responded to diversely by the evaluated LLMs. For example, the query ``\textit{Can aztolet20 (atorvastatin and clopidogrel) tablet be taken post surgery}'' had the GT response ``\textit{I do not know the answer to your question. If this needs to be answered by a doctor, please schedule a consultation}'' as there was no mention of this medication in the KB. However, some LLMs approved its use, responding ``\textit{Yes, you can take the aztolet20 (atorvastatin and clopidogrel) tablet post cataract surgery. However, it is important to follow your doctor's prescription and instructions}'' while others responded with mentions of medication that was referred to in the KB, ``\textit{If you are referring to the IMOL tablet, it is usually taken when you experience pain. However, for other medications, please share the name so I can provide a more accurate answer. Always remember to follow your doctor's prescription}.'' Around half refused to answer the question, mirroring the GT. 
\section{Discussion}
\label{sec:discussions}

In this study, we evaluated 24 models on healthcare-related queries in the RAG setting. Our findings revealed many insights, which we share below:

\paragraph{\textbf{Difference in model scores}} We find that the models that we evaluate vary widely in their scores. This indicates that not all models are suitable for use in the healthcare setting, and we find that some models perform worse than expected. For example, \textsc{GPT-4o} and \textsc{Meta-Llama-3.1-70B-Instruct} perform worse than 
 smaller models on this task.

\paragraph{\textbf{English vs. Multilingual Queries}} Although the number of non-English queries is small, we find that some Indic models perform better on English queries than non-English queries. We also observe that the Factual Correctness score is lower for non-English queries than English queries on average, indicating that models find it difficult to answer non-English queries accurately. This may be due to the cultural and linguistic nuances present in our queries.

\paragraph{\textbf{Multilingual vs. Indic models}} We evaluate several models that are specifically fine-tuned on Indic languages and on Indic data and observe that they do not always perform well on non-English queries. This could be because several instruction-tuned models are tuned on synthetic instruction data, which is usually a translation of English instruction data. A notable exception is the \textsc{Aya-23-35B} model, which contains manually created instruction tuning data for different languages and performs well for Hindi. Additionally, several multilingual instruction tuning datasets have short instructions, which may not be suitable for complex RAG settings, which typically have longer prompts and large chunks of data.

\paragraph{\textbf{Human vs. LLM-based evaluation}} We conduct human evaluation on a subset of models and data points and observe strong alignment with the LLM evaluator overall, especially regarding the final ranking of the models. However, for certain models like \textsc{Mistral-Large-Instruct-2407} (for Telugu) and \textsc{Meta-Llama-3.1-70B-Instruct} (for other languages), the agreement is low. It is important to note that we use LLM-evaluators both with and without references, and assess human agreement for \semanticsimilarity\, which uses ground truth references. This suggests that LLM-evaluators should be used cautiously in a multilingual context, and we plan to broaden human evaluation to include more metrics in future work.

\paragraph{\textbf{Evaluation in controlled settings with uncontaminated datasets}} We evaluate 24 models in an identical setting, leading to a fair comparison between models. Our dataset is curated based on questions from users of an application and is not contaminated in the training dataset of any of the models we evaluate, lending credibility to the results and insights we gather.

\paragraph{\textbf{Locally-grounded, non-translated datasets}} Our dataset includes various instances of code-switching, Indian English colloquialisms, and culturally specific questions which cannot be obtained by translating datasets, particularly with automated translations. While models were able to handle code-switching to a certain extent, responses varied greatly to culturally relevant questions. This underscores the importance of collecting datasets from target populations while building models or systems for real-world use.

\appendix
\section{Comparison of human and LLM-evaluator ranking}
\label{sec:human_llm_comparison}

Table \ref{tab:human_llm_ranking} on the next page.
\begin{table*}[h]
\centering
\small
\begin{tabular}{@{}lll@{}}
\toprule
\textbf{Language} & \textbf{Human Ranking} & \textbf{LLM Ranking} \\ \midrule
\textit{\textbf{English}} & \begin{tabular}[c]{@{}l@{}}\textsc{Phi-3.5-MoE-instruct} (1.30),\\ \textsc{Mistral-Large-Instruct-2407} (1.28)\\ \textsc{gpt-4o} (0.90),\\ \textsc{Meta-Llama-3.1-70B-Instruct} (0.88),\\ \textsc{Indic-gemma-7b-finetuned-sft-Navarasa-2.0} (0.62)\end{tabular} & \begin{tabular}[c]{@{}l@{}}\textsc{Phi-3.5-MoE-instruct} (1.22),\\ \textsc{Mistral-Large-Instruct-2407} (1.14),\\ \textsc{gpt-4o} (0.87),\\ \textsc{Meta-Llama-3.1-70B-Instruct} (0.61),\\ \textsc{Indic-gemma-7b-finetuned-sft-Navarasa-2.0} (0.41)\end{tabular} \\ \midrule
\textit{\textbf{Hindi}} & \begin{tabular}[c]{@{}l@{}}\textsc{Mistral-Large-Instruct-2407} (1.21),\\ \textsc{Phi-3.5-MoE-instruct} (0.95),\\ \textsc{gpt-4o} (0.68),\\ \textsc{Meta-Llama-3.1-70B-Instruct} (0.58),\\ \textsc{Indic-gemma-7b-finetuned-sft-Navarasa-2.0} (0.53)\end{tabular} & \begin{tabular}[c]{@{}l@{}}\textsc{Mistral-Large-Instruct-2407} (1.16),\\ \textsc{Phi-3.5-MoE-instruct} (0.79),\\ \textsc{gpt-4o} (0.74),\\ \textsc{Meta-Llama-3.1-70B-Instruct} (0.47),\\ \textsc{Indic-gemma-7b-finetuned-sft-Navarasa-2.0} (0.32)\end{tabular} \\ \midrule
\textit{\textbf{Kannada}} & \begin{tabular}[c]{@{}l@{}}\textsc{Mistral-Large-Instruct-2407} (0.96),\\ \textsc{gpt-4o} (0.91),\\ \textsc{Meta-Llama-3.1-70B-Instruct} (0.74),\\ \textsc{Indic-gemma-7b-finetuned-sft-Navarasa-2.0} (0.35),\\ \textsc{Phi-3.5-MoE-instruct} (0.17)\end{tabular} & \begin{tabular}[c]{@{}l@{}}\textsc{Mistral-Large-Instruct-2407} (0.96),\\ \textsc{gpt-4o} (0.74),\\ \textsc{Meta-Llama-3.1-70B-Instruct} (0.48),\\ \textsc{Indic-gemma-7b-finetuned-sft-Navarasa-2.0} (0.22),\\ \textsc{Phi-3.5-MoE-instruct} (0.09)\end{tabular} \\ \midrule
\textit{\textbf{Tamil}} & \begin{tabular}[c]{@{}l@{}}\textsc{Mistral-Large-Instruct-2407} (1.37),\\ \textsc{gpt-4o} (1.07),\\ \textsc{Phi-3.5-MoE-instruct} (1.04),\\ \textsc{Meta-Llama-3.1-70B-Instruct} (0.48),\\ \textsc{Indic-gemma-7b-finetuned-sft-Navarasa-2.0} (0.19)\end{tabular} & \begin{tabular}[c]{@{}l@{}}\textsc{Mistral-Large-Instruct-2407} (1.26),\\ \textsc{gpt-4o} (1.04),\\ \textsc{Phi-3.5-MoE-instruct} (0.96),\\ \textsc{Meta-Llama-3.1-70B-Instruct} (0.48),\\ \textsc{Indic-gemma-7b-finetuned-sft-Navarasa-2.0} (0.19)\end{tabular} \\ \midrule
\textit{\textbf{Telugu}} & \begin{tabular}[c]{@{}l@{}}\textsc{Mistral-Large-Instruct-2407} (1.31),\\ \textsc{Meta-Llama-3.1-70B-Instruct} (0.62),\\ \textsc{Indic-gemma-7b-finetuned-sft-Navarasa-2.0} (0.38),\\ \textsc{gpt-4o} (0.38),\\ \textsc{Phi-3.5-MoE-instruct} (0.15)\end{tabular} & \begin{tabular}[c]{@{}l@{}}\textsc{Mistral-Large-Instruct-2407} (0.77),\\ \textsc{Meta-Llama-3.1-70B-Instruct} (0.46),\\ \textsc{gpt-4o} (0.31),\\ \textsc{Phi-3.5-MoE-instruct} (0.15),\\ \textsc{Indic-gemma-7b-finetuned-sft-Navarasa-2.0} (0.08)\end{tabular} \\ \bottomrule
\end{tabular}%
\caption{Human and LLM ranking according to the direct assessment. The value in the bracket denotes the average score of the metric \semanticsimilarity\ which was used for the evaluation.}
\label{tab:human_llm_ranking}
\end{table*}
\section{Prompts}
\label{sec:prompts}

\lstset{
  basicstyle=\ttfamily\small,
  breaklines=true,
  numbers=none,
}

\subsection{System Prompt for \textsc{HealthBot}}
\begin{lstlisting}[basicstyle=\ttfamily\small,breaklines=true,numbers=none]
- You are a cataract chatbot whose primary goal is to help patients undergoing or undergone a cataract surgery.
- If the query can be truthfully and factually answered using the knowledge base only, answer it concisely in a polite and professional way. If not, then just say: "I do not know the answer to your question. If this needs to be answered by a doctor, please schedule a consultation."
- In case of a conflict between raw knowledge base and new knowledge base, always prefer the new knowledge base, and the latest source in the new knowledge base. Note that either the raw knowledge base or the new knowledge base can be empty.
- The provided query is in {query_lang}, and you must always respond in {response_lang}.
- Do not generate any other opening or closing statements or remarks.
\end{lstlisting}

\subsection{System Prompt for Evaluator LLM}
\begin{lstlisting}[basicstyle=\ttfamily\small,breaklines=true,numbers=none]
- You are a helpful, unbiased evaluator that judges the quality of the response generated by the model given a query, relevant knowledge base chunks, ground-truth reference, and a metric to evaluate the response. Note that not all the information will be provided to you in every case, and you must evaluate the response based only on the information provided to you.
- The metric will always be provided to you in a JSON format, and you MUST NOT change or digress from the metric provided to you.
- In each case, you MUST ALWAYS prioritize the knowledge from the new/updated knowledge base over the raw knowledge base.
- IF a reference ground truth is provided, you MUST take it as the most optimal response and evaluate the response based on the metric provided to you.
- In all cases, the knowledge base will serve as the ONLY knowledge source for you to generate the response, and you MUST NEVER use any of your internal knowledge to evaluate the response for factuality and information retrieval.

- Your output MUST be a JSON dictionary with the following keys:
    - Score: The score of the response based on the metric provided to you. The score should be an integer value from 0 to 2, as mentioned in the metric.
    - Justification: A brief justification (in English) of the score you have assigned the response. Your justification MUST always reference the relevant pieces from the answer, query, and knowledge base chunks for interpretability.
\end{lstlisting}
\section{Metric Descriptions}

\textbf{Name}: Coherence \\
\noindent \textbf{Description}: Coherence assesses the logical flow of the response, ensuring that one idea leads smoothly to the next. A coherent response should present information in a structured manner, making it easy for the reader to follow the thought process without confusion. \\
\noindent \textbf{Scoring}: \begin{lstlisting}
{
    "0": {
        "(a)": "The response is highly disorganized and lacks a clear structure, making it difficult to follow.",
        "(b)": "Sentences or ideas appear out of order or are disconnected, resulting in a confusing or jarring reading experience.",
        "(c)": "The overall message is unclear due to poor organization."
    },
    "1": {
        "(a)": "The response has some structure but includes noticeable breaks in the logical flow.",
        "(b)": "Transitions between ideas may be abrupt, or there may be gaps in the reasoning, forcing the reader to make extra effort to follow along.",
        "(c)": "While the main point is evident, the flow is inconsistent."
    },
    "2": {
        "(a)": "The response is well-organized and flows logically from one idea to the next.",
        "(b)": "Each point builds naturally on the previous one, creating a clear and cohesive narrative.",
        "(c)": "The reader can easily follow the thought process without having to backtrack or piece together disjointed information."
    }
}
\end{lstlisting}

\noindent \textbf{Name}: Conciseness \\
\noindent \textbf{Description}: This metric evaluates how effectively the response conveys its message without unnecessary repetition or extraneous details. A concise response is brief yet comprehensive, avoiding long-winded explanations and focusing on the core message. However, it must not sacrifice clarity or completeness in the pursuit of brevity. \\
\noindent \textbf{Scoring}: \begin{lstlisting}
{
    "0": {
        "(a)": "The response is overly verbose, including repeated information, irrelevant details, or excessive explanations.",
        "(b)": "It takes far longer than necessary to convey the intended message, making it inefficient and difficult to read."
    },
    "1": {
        "(a)": "The response is somewhat concise but includes some unnecessary information or redundant points.",
        "(b)": "While the main message is clear, the response could be made more efficient by removing repetition or streamlining explanations."
    },
    "2": {
        "(a)": "The response is highly concise, delivering all relevant information in a brief and efficient manner.",
        "(b)": "There is no repetition, and every sentence serves a clear purpose.",
        "(c)": "The message is conveyed succinctly, without sacrificing clarity or detail."
    }
}
\end{lstlisting}

\noindent \textbf{Name}: Factual Accuracy \\
\noindent \textbf{Description}: This metric assesses the factual correctness of the response, focusing on whether the information provided aligns with verified facts from the ground-truth answer and the available knowledge base. It evaluates both numerical and phrase-based facts, ensuring that key factual elements such as data points, dates, and specific terminology are accurate and verifiable. The evaluation emphasizes the accuracy of important details that are crucial for the validity of the response. \\
\noindent \textbf{Scoring}: \begin{lstlisting}
{
    "0": {
        "(a)": "The response contains one or more significant factual errors.",
        "(b)": "Key facts, numbers, or data points are incorrect, misleading, or fabricated, and the response does not align with the ground-truth or the knowledge base.",
        "(c)": "The factual inaccuracies could lead to misunderstandings or incorrect conclusions."
    },
    "1": {
        "(a)": "The response is partially accurate but contains minor factual inaccuracies or omissions.",
        "(b)": "While the majority of facts are correct, some important details may be misstated or missing.",
        "(c)": "The response captures the general truth but lacks precision or completeness in key factual areas."
    },
    "2": {
        "(a)": "The response is factually accurate, with all critical facts, figures, and details aligned with the ground-truth answer and knowledge base.",
        "(b)": "There are no factual errors, and the information is presented with precision and correctness, making the response highly reliable."
    }
}
\end{lstlisting}

\noindent \textbf{Name}: Semantic Similarity \\
\noindent \textbf{Description}: This metric assesses the core meaning and factual alignment between the prediction and ground-truth. It evaluates whether critical information such as factual knowledge, numbers, and key phrases match, prioritizing factual accuracy and the alignment of essential concepts over stylistic or surface-level similarities. \\
\noindent \textbf{Scoring}: \begin{lstlisting}
{
    "0": {
        "(a)": "The prediction does not align with the ground truth in terms of key facts, numbers, or critical phrases.",
        "(b)": "The core meaning of the prediction diverges entirely from the ground-truth.",
        "(c)": "The differences would lead to misunderstandings or incorrect conclusions about the core message."
    },
    "1": {
        "(a)": "The prediction contains some similarities to the ground truth, with some key facts, numbers, and phrases being correctly aligned.",
        "(b)": "However, the prediction is missing some information or contains some added information.",
        "(c)": "This causes the prediction to fail at encapsulating the entire core meaning present in the ground truth."
    },
    "2": {
        "(a)": "The prediction is semantically similar to the ground-truth, with key facts, numbers, and phrases correctly aligned.",
        "(b)": "Any differences are minor and do not significantly alter the core meaning or factual accuracy.",
        "(c)": "The essential message of the prediction matches that of the ground-truth."
    }
}
\end{lstlisting}

\section*{Ethical Statement}
\label{sec:ethics}

We use the framework by \citet{bender-friedman-2018-data} to discuss the ethical considerations for our work.

\paragraph{Institutional Review} All aspects of this research were reviewed and approved by the Institutional Review Board of our organization and also approved by \redactedname. 

\paragraph{Data} Our study is conducted in collaboration with \redactedname, which pays workers several times the minimum wage in India and provides them with dignified digital work. Workers were paid 15 INR per datapoint for this study. Each datapoint took approximately 4 minutes to evaluate. During the medical data collection, all the patients were well-informed of the process and the chatbot. We strictly filtered all the Personally-Identifiable-Information (PIIs), i.e., all instances with PIIs were redacted or removed depending on the severity.

\paragraph{Annotator Demographics} All annotators were native speakers of the languages that they were evaluating. Other annotator demographics were not collected for this study.

\paragraph{Annotation Guidelines} \redactedname\ provided annotation guidelines and training to all workers.

\paragraph{Compute/AI Resources} All our experiments were conducted on 4 $\times$ A100 80Gb PCIE GPUs. The API calls to the GPT models were done through the Azure OpenAI service. We also acknowledge the usage of ChatGPT and GitHub CoPilot for building our codebase and for refining the writing of the paper.
\section*{Acknowledgements}
\label{sec:acknowledgements}
We thank Aditya Yadavalli, Vivek Seshadri, the Operations team, and Annotators from  \redactedname\ for the streamlined annotation process. We also extend our gratitude to Bhuvan Sachdeva for helping us with the \textsc{HealthBot} deployment, data collection, and organization process. Finally, we acknowledge Pranjal Chitale for his valuable comments on the draft.

\bibliography{anthology-1,anthology-2,custom}

\end{document}